\documentclass[letterpaper, 10 pt, conference]{ieeeconf} 
\usepackage[letterpaper, left=48pt, right=60pt, bottom=48pt, top=55pt]{geometry}

\IEEEoverridecommandlockouts 
\title{\LARGE \bf
CDE: Concept-Driven Exploration for Reinforcement Learning
}

\author{Anonymous Authors}
\author{Le Mao$^{1}$, Andrew H. Liu$^{2}$, Renos Zabounidis$^{3}$, Yanan Niu$^{4}$, Zachary Kingston$^{2}$, Joseph Campbell$^{2}$%
\thanks{$^{1}$ LM is with the  Department of Electrical and Computer Engineering, Purdue University, West Lafayette, IN 47906, USA
        {\tt\small mao214@purdue.edu}}%
\thanks{$^{2}$ AL, ZK and JC are with the Department of Computer Science, Purdue University, West Lafayette, IN 47906, USA
        {\tt\small \{liu3458, zkingston, joecamp\}@purdue.edu}}%
\thanks{$^{3}$ RZ is with the Robotics Institute, Carnegie Mellon University, Pittsburgh, PA 15213, USA
        {\tt\small renosz@andrew.cmu.edu}}%
\thanks{$^{4}$ YN is with the Department of Management of Technology, EPFL, Lausanne, Switzerland
        {\tt\small yanan.niu@epfl.ch}}
}

\usepackage[utf8]{inputenc}
\usepackage[T1]{fontenc}
\usepackage[english]{babel}
\usepackage{xurl}
\usepackage{graphicx}
\usepackage[absolute,overlay]{textpos}
\usepackage{caption}
\usepackage{amssymb}
\usepackage{algorithm}
\usepackage{algpseudocode}
\usepackage{amsmath}
\usepackage{setspace}
\usepackage{multirow}
\usepackage{siunitx} 
\usepackage{pifont}  
\usepackage{graphicx}
\usepackage{booktabs}
\usepackage{hhline}
\usepackage{subcaption}
\usepackage[x11names]{xcolor}
\usepackage{mdframed}

\usepackage[english=american]{csquotes}
\MakeOuterQuote{"}

\makeatletter
\let\NAT@parse\undefined
\makeatother

\usepackage[numbers,sort&compress]{natbib}

\usepackage[pdfa,colorlinks,bookmarksopen,bookmarksnumbered,allcolors=linkcolor]{hyperref}

\usepackage[nameinlink,capitalise]{cleveref}
\crefname{table}{Tbl.}{Tbls.}
\Crefname{table}{Tbl.}{Tbls.}
\crefname{figure}{Fig.}{Figs.}
\Crefname{figure}{Fig.}{Figs.}
\crefname{equation}{Eq.}{Eqs.}
\Crefname{equation}{Eq.}{Eqs.}
\crefname{algorithm}{Alg.}{Algs.}
\Crefname{algorithm}{Alg.}{Algs.}
\crefformat{section}{#2Sec.~#1#3}

\definecolor{purduegold}{HTML}{C28E0E} %
\definecolor{linkcolor}{HTML}{2b3b5e}

\hypersetup{
bookmarksopen,
bookmarksnumbered,
colorlinks=true,
allcolors=purduegold,
}
\definecolor{hicell}{HTML}{ebd99f}  %

\newcommand{\hibf}[1]{\textcolor{purduegold}{\bf#1}}

\usepackage{flushend}

\begin{document}

\maketitle
\thispagestyle{empty}
\pagestyle{empty}

%%%%%%%%%%%%%%%%%%%%%%%%%%%%%%%%%%%%%%%%%%%%%%%%%%%%%%%%%%%%%%%%%%%%%%%%%%%%%%%%
\begin{abstract}
Intelligent exploration remains a critical challenge in reinforcement learning (RL), especially in visual control tasks.
Unlike low-dimensional state-based RL, visual RL must extract task-relevant structure from raw pixels, making exploration inefficient.
We propose Concept-Driven Exploration (CDE), which leverages a pre-trained vision–language model (VLM) to generate object-centric visual concepts from textual task descriptions as weak, potentially noisy supervisory signals.
Rather than directly conditioning on these noisy signals, CDE trains a policy to reconstruct the concepts via an auxiliary objective, learning general representations of the concepts and using reconstruction accuracy as an intrinsic reward to guide exploration toward task-relevant objects.Across five challenging simulated visual manipulation tasks, CDE achieves efficient, targeted exploration and remains robust to both synthetic errors and noisy VLM predictions.
Finally, we demonstrate real-world transfer by deploying CDE on a Franka arm, attaining an 80\% success rate in a real-world manipulation task.
Code and videos are available at: \url{https://sites.google.com/view/concept-learn/home}.

\end{abstract}
\section{Introduction}
Reinforcement learning (RL) has shown impressive performance across a variety of robotic tasks, including manipulation~\cite{RL_Manipulation1,RL_Manipulation2,RL_Manipulation3}, navigation~\cite{RL_Navigation1, RL_Navigation2, RL_Navigation3} and task planning~\cite{RL_Task_Planning1}. Yet exploration remains challenging, especially under sparse or delayed rewards where random exploration leads to wasteful environment interactions.
This difficulty is amplified in visual control: policies must first learn to extract task-relevant objects and relations from high-dimensional images in order to effectively ground credit assignment. Recent works have explored leveraging pre-trained VLMs to automatically generate dense reward signals, incorporating task-related domain knowledge to facilitate learning in complex tasks~\cite{wang2024rl, sontakke2023roboclip, mahmoudieh2022zero, rocamonde2023vision, adeniji2024language, ma2023eureka}. 
While promising, such approaches rely on point estimates from VLMs that can produce noisy or inaccurate outputs in practice. Directly optimizing over these imperfect signals can misguide exploration and reduce training effectiveness.

\begin{figure}
    \centering
    \includegraphics[width=1.\linewidth]{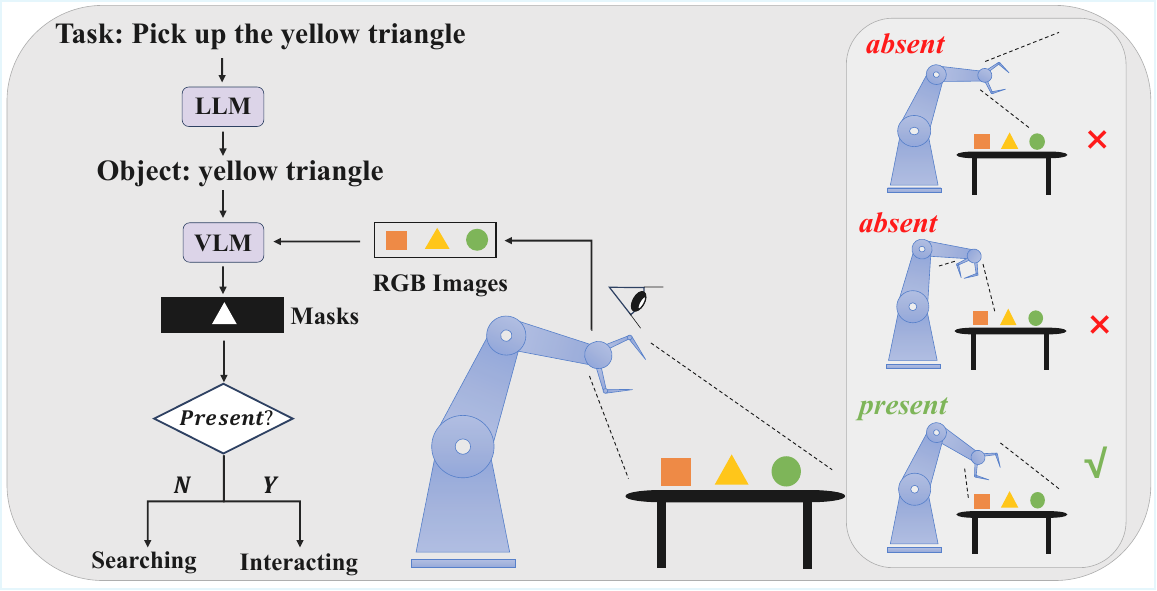}
    \caption{Concept-Driven Exploration overview. Task-relevant objects are identified from a task description and then used by a VLM to generate segmentation masks for each object.
    These segmentation masks shape policy representation learning and guide exploration.}
    \label{fig:Overview}
\end{figure}

This raises a fundamental question: how can we leverage semantic guidance from VLMs while remaining robust to their inherent noise?
In this work, we propose Concept-Driven Exploration (CDE), a method for robust and sample-efficient policy learning under noisy VLM guidance.
Instead of directly conditioning on VLM-generated rewards, CDE takes a representation-first approach.
From a natural language task description, a VLM proposes relevant visual concepts---concrete, task-level cues such as the segmentation mask of a target object (see~\cref{fig:Overview}).
Rather than treating these concepts as ground-truth supervision~\cite{huang2025roboground, stone2023open, liu2024moka}, CDE instead assumes the visual concepts are inherently noisy and treats them as weakly supervised learning targets~\cite{jaderberg2016reinforcement, mirowski2016learning}.
The policy is trained to predict these concepts via an auxiliary reconstruction loss, and the resulting reconstruction error is used as an intrinsic reward to guide exploration. Intuitively, the VLM-generated concepts serve as ``hints'' which help the policy learn to recognize task-relevant objects, while the intrinsic reward guides exploration toward them.

The design yields three key benefits. First, exploration becomes object-centric, encouraging the agent to attend to the task-relevant regions instead of background distractors. Second, by modeling VLM outputs as weak supervision rather than oracle rewards, training remains stable even when semantic predictions are imperfect~\cite{rolnick2017deep}. Third, once concept prediction is learned, the VLM is no longer required.

Beyond noisy semantic supervision, practical robotic systems often operate under partial visual observability. While many prior works assume fixed global or multi-camera views, such assumptions may not hold at deployment. We therefore study the more challenging setting where only wrist-mounted observations are available. Compared to a fixed global camera that offers relatively stable visual observations, a wrist-mounted camera returns frames with drastic visual changes, and the target object is not always visible in the camera view, making policy learning more challenging.
To address this, we integrate Concept Embedding Models (CEMs) to learn dual object representations conditioned on object visibility.
Since policy behavior is visibility-dependent---interact with the object if it is visible, search for it otherwise---the two representations capture complementary features, resulting in improved learning efficiency.

We summarize our contributions as follows:

\begin{itemize}
    \item We propose a concept-driven exploration method that utilizes VLMs to generate visual concepts in a zero-shot manner, with no manual annotations.

    \item We treat these visual concepts as weakly supervised targets to learn task-relevant representations and generate intrinsic rewards for exploration.

    \item We integrate CEMs with the policy network to represent both the presence and absence of task-relevant objects, resulting in representations compatible with wrist-mounted cameras where objects may not be visible.

    \item We empirically show that CDE outperforms baselines on five visual manipulation tasks and evaluate transfer to real-world settings.

\end{itemize}

\begin{figure*}
    \centering \vspace{3mm}
    \includegraphics[width=0.98\linewidth]{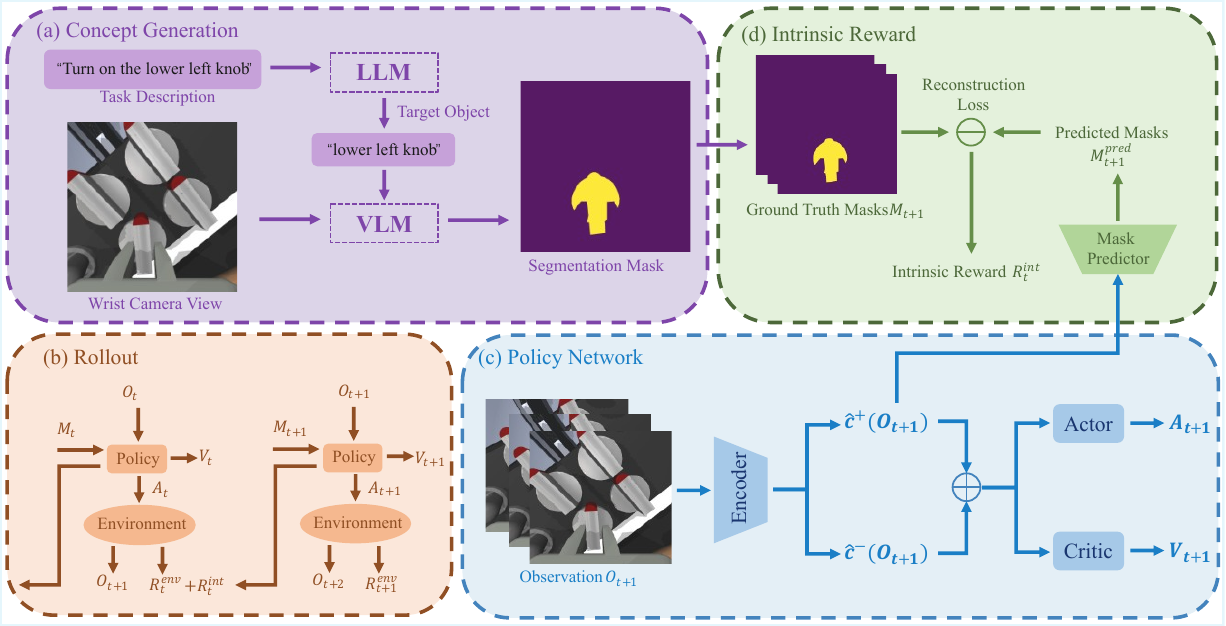}
    \caption{Architecture. (a) The LLM parses the task description to extract the target object. The VLM segments masks on input RGB images. (b) During training, the policy takes segmentation masks as additional input and generates intrinsic reward signals for the last timestep. (c) At each timestep $t+1$, the policy network receives environment observation $O_{t+1}$, and encodes $O_{t+1}$ into a positive embedding $\hat{\textbf{c}}^{+}(O_{t+1})$ and a negative embedding $\hat{\textbf{c}}^{-}(O_{t+1})$, the final concept embedding is a weighted sum of the two embeddings. (d) The segmentation masks $M_{t+1}$ are used to supervise the mask reconstruction from the positive embedding $\hat{\textbf{c}}^{+}(O_{t+1})$ and generate intrinsic reward signal $R_{t}^{\text{int}}$.}
    \label{fig:Architecture}
\end{figure*}

\section{Related Work}
\subsection{Vision-based Reinforcement Learning}

Existing works have explored both model-based and model-free RL approaches for visuomotor control. For model-based RL, PlaNet~\cite{PlaNet} learns environment dynamics from only image observations. Visual Foresight~\cite{visual_foresight} predicts future frames conditioned on past actions and current image observation. Without access to an explicit dynamics model~\cite{model-based_RL}, model-free approaches are typically less efficient than model-based approaches. Previous works have tried different methods to improve sample efficiency of model-free RL in image space. CURL~\cite{CURL} leverages contrastive learning to learn rich representations and thus accelerate policy learning. DrQ~\cite{DrQ} and DrQv2~\cite{DrQv2} apply image augmentation to the training process and achieve state-of-the-art performance on several DeepMindControl tasks. SEAR~\cite{SEAR-RL} learns agent and background representations through reconstruction. DEAR~\cite{DEAR} extends SEAR by maximizing distance between two representations but without background reconstruction. The above methods don't explicitly model task-relevant objects, potentially leading exploration to rely on global image features that include background variations and irrelevant signals. In contrast, CDE learns task-relevant visual concepts of target objects to promote more focused and efficient exploration.

\subsection{Concept Learning in Reinforcement Learning}
 
Previous works have explored reconstructing pre-defined concepts from intermediate embeddings and using these for downstream policy learning. Concept Policy Models ~\cite{concept_learning_for_MARL} apply Concept Bottleneck Models (CBM)~\cite{CBM} to multi-agent RL. SCoBots~\cite{ScoBots} extracts object-related symbolic concepts from raw observations and refines them into human-understandable relational concepts, with concept pruning and reward shaping performed by human experts. LICORICE~\cite{LICORICE} focuses on reducing human labor in labeling concepts and querying a VLM for labels. However, the aforementioned works require pre-defined semantic concepts,which may vary across environments. CED addresses these limitations through a two-fold design. first, it adopts domain-agnostic segmentation masks as concepts, improving  generalization across environments. Second, instead of directly conditioning on potentially inaccurate VLM outputs, CDE treats them as noisy supervision and enables more robust policy learning.

\subsection{Intelligent Exploration}

Intelligent and efficient exploration remains a critical challenge in RL.  For example, frequency-based methods encourage novelty-seeking behavior by giving higher rewards to states with lower visitation counts~\cite{explorationRL_survey}. Curiosity-based methods incentivize the policy to visit unseen states through intrinsic motivation by predicting future states and using the prediction error as an additional reward. ICM~\cite{ICM} predicts environment dynamics with inverse and forward models. RND~\cite{RND} and variants such as DRND~\cite{DRND} force the prediction network to approximate a randomly initialized network. VIME~\cite{VIME} leverages information gained from the agent's belief. These approaches primarily reward state novelty without explicitly considering task relevance. In object-centric tasks, this may lead to exploration that is diverted toward background variations rather than interactions with the target object.

\section{Preliminaries}
\label{sec:preliminaries}
\subsection{Reinforcement Learning} Reinforcement learning is modeled as a Markov decision process (MDP) described by the tuple $ \langle \mathcal{S}, \mathcal{A}, P, R, \gamma  \rangle$. $\mathcal{S}$ denotes the state space, $\mathcal{A}$ denotes the action space. $P(s_{t+1} | s_{t}, a_{t}) \in [0, 1]$ denotes the transition probability from state $s_{t}$ to $s_{t+1}$ given the action $a_{t}$. $R(s_{t}, a_{t}) \in \mathbb{R}$ denotes the reward function. $\gamma \in (0, 1)$ is the discount factor. At each time step $t$, the agent takes an action $a_{t}$ from policy $ \pi (a_{t}|s_{t})$ given the state $s_{t}$. The objective is to maximize the expected return $ \mathbb{E}_{s, a} \left[\sum_{t=0}^{\infty} \gamma^{t} R(s_{t}, a_{t}) \right]$.

\subsection{Concept Embedding Models}
\label{subsec:concept embedding model}
Concept Embedding Models (CEMs)~\cite{cem} were originally proposed as interpretable image classification models. CEMs map the input image $\textbf{x}$ into a set of human-defined concepts $ C = \{c_{1}, c_{2}, \cdots ,c_{n}\}$ through an intermediate layer $f(\cdot)$. A concept $c_{i}$ is represented by two embeddings $\hat{\textbf{c}}_{i}^{+}$ and $\hat{\textbf{c}}_{i}^{-}$:
\begin{equation} 
    \hat{\textbf{c}}_{i}^{+},\hat{\textbf{c}}_{i}^{-} = f(\textbf{x}).
\end{equation}

The positive embedding $\hat{\textbf{c}}_{i}^{+}$ represents that the concept $c_{i}$ is active, while the negative embedding $\hat{\textbf{c}}_{i}^{-}$ represents that the concept is inactive. A probability $\hat{p}_{i} \in [0, 1]$ is predicted from the two embeddings through a probability generator $P_{i}(\cdot)$ to indicate whether the concept is active. The final concept embedding $ \hat{\textbf{c}}_{i} $ is the weighted mixture of $\hat{\textbf{c}}_{i}^{+}$ and $\hat{\textbf{c}}_{i}^{-}$ which is formulated as:
\begin{equation}
\label{equation 2}
    \hat{\textbf{c}}_{i} = \hat{p}_{i} \hat{\textbf{c}}_{i}^{+} + (1 - \hat{p}_{i}) \hat{\textbf{c}}_{i}^{-} \ \ \text{where} \ \ \hat{p}_{i} = P (\hat{\textbf{c}}_{i}^{+},  \hat{\textbf{c}}_{i}^{-}).
\end{equation}

\section{Concept-Driven Exploration}
We propose CDE, a concept-driven exploration framework for robust and sample-efficient policy learning in vision-based manipulation tasks (See \cref{fig:Architecture}).
Before each episode, an LLM extracts relevant target objects from a task description.
During policy rollout, a VLM provides per-step (potentially noisy) segmentation masks of each object.
The policy encodes each image into a concept embedding used by downstream policy layers.
We train this embedding jointly with the RL objective and a mask-reconstruction loss, which is computed by decoding the target mask from the embedding and comparing it to the VLM-generated mask.
The reconstruction loss serves two roles: it acts as an intrinsic reward to encourage object-centric exploration, and as an auxiliary optimization objective that shapes the representations toward object-relevant features (See \cref{alg:Pseudocode}).
Importantly, rather than feeding segmentation masks directly as observations, we treat them as supervision for the policy's internal representations.

\subsection{Concept Generation}
\label{subsec:concept generation}
Prior works~\cite{LICORICE, ScoBots,concept_learning_for_MARL} often rely on human-interpretable concepts (e.g., relative position, orientation), however, in visual control tasks such concepts are difficult to define purely from RGB images and task descriptions. To solve this problem, we propose to use the \textit{segmentation mask} of the target object as a concept. We first use an LLM to generate a list of objects that the policy should interact with, allowing object identification to generalize across tasks without manual specification. Below is an example for generating a list of objects for the \textit{Kitchen-Microwave} task using OpenAI GPT-4~\cite{GPT4}:

\begin{samepage}
\begin{mdframed}
  \noindent \textbf{Prompt:} Task description: "Open the microwave door". Give me a list of objects to interact with in order to solve the task. Please return a Python list. Do not output anything else.

  \noindent \textbf{Output:} ["microwave door handle"]
\end{mdframed}
\end{samepage}
Given the resulting task-related objects, we query a VLM to generate segmentation masks for each object in the list from visual observations.

\subsection{Concept Learning with Concept Embedding Models}
Because VLM-generated segmentation masks are imperfect, directly feeding them as policy inputs leads to unstable learning. Instead, we use the masks as auxiliary targets to encourage object-centric representations.
Unlike previous works~\cite{SEAR-RL, DEAR} that use a global camera, our wrist-mounted camera frequently fails to capture the target object. In such frames, embeddings that encode only object-specific features may not aid learning (See \cref{sec:ablation studies}).
To address this partial observability, we adopt CEMs that associate each concept $i$ with two embeddings, 
$\hat{\mathbf{c}}_i^{+}$ and $\hat{\mathbf{c}}_i^{-}$, representing the object-present and object-absent regimes, respectively. We combine the two embeddings via a gated formulation for the downstream policy (\cref{equation 2}):
\begin{equation}
    \hat{\textbf{c}}_{i} = p_{i} \hat{\textbf{c}}_{i}^{+} + (1 - p_{i}) \hat{\textbf{c}}_{i}^{-} \ \ \text{where} \ \ p_{i} = \begin{cases}
    1 & \text{if px} \geq \epsilon \\
    0 & \text{if px} < \epsilon,
    \end{cases}  
\end{equation}
where $p_i$ indicates whether the target object is present. Unlike the original CEM formulation, we do not predict the $p_i$ from the embeddings, but instead derive it directly from the segmentation mask by thresholding the number of active mask pixels $px$ with a small threshold $\epsilon$ (e.g., 20). During training, $p_i$ is computed from the VLM-generated mask; at deployment, it is obtained from the predicted mask.

The Q-function regression loss $\mathcal{L}_{\text{critic}}$ is computed following DrQv2~\cite{DrQv2}. Additionally, we introduce an auxiliary reconstruction loss $\mathcal{L}_{\text{recons}}$. Specifically, we use the positive embedding to reconstruct the segmentation mask of the target object. This encourages the positive embedding to encode object-related visual information. The final objective is:
\begin{equation}
    \mathcal{L}_{\text{total}} = \alpha \mathcal{L}_{\text{critic}} + \beta \mathcal{L}_{\text{recons}}.
\end{equation}

\subsection{Intrinsic Reward}
In addition to learning object-related representations, we find that segmentation mask reconstruction can also be used to generate intrinsic reward signals to incentivize exploration. 

\begin{equation}
    {r}^{\text{total}}_{t} = {r}^{\text{env}}_{t} + \gamma \text{clip} \left(\mathcal{L}_{\text{recons}}, 0, 1 \right).
\end{equation}
Since the model is supervised to reconstruct segmentation masks from the positive embedding $\hat{\textbf{c}}_{i}^{+}$ through a mask predictor $MP_{\phi}$, the reconstruction loss is expected to be lower for previously visited states than for unseen states. Therefore, the policy is encouraged to visit novel states where the target object is present to maximize the reconstruction loss.

\begin{algorithm}[t]

\caption{Concept-Driven Exploration}
\label{alg:Pseudocode}
\begin{spacing}{1.05}
\begin{algorithmic}[1]
\Require LLM, VLM, Encoder $E_{\theta}$, Mask Predictor $MP_{\phi}$, task description $ d $

\State $\text{Object} \ C\gets \text{Prompt}\left(\text{LLM}, d\right)$

\For{$t = 1$ to $T$}

\State $\text{Collect transition} \left(o_{t}, m_{t}^{\text{gt}}, a_{t}, r_{t}^{\text{env}}, o_{t+1}\right)$

\State $m_{t+1}^{\text{gt}}\gets \text{VLM}\left(o_{t+1}, C\right)$

\State $\hat{c}^{+}_{t+1}, \hat{c}^{-}_{t+1} \gets E_{\theta}\left(o_{t+1}\right)$

\State $ m_{t+1}^{\text{pred}} \gets MP_{\phi}\left(\hat{c}^{+}_{t+1}\right)$ \Comment{Mask Prediction}

\State $\mathcal{L}_{\text{recons}} = \mathcal{L}_{\text{BCE}}\left(m_{t+1}^{\text{pred}}, m_{t+1}^{\text{gt}}\right)$ 

\State $r_{t}^{\text{int}} = \gamma \text{clip}\left(\mathcal{L}_{\text{recons}}, 0, 1\right)$ \Comment{Intrinsic Reward}

\State $\mathcal{D} \gets \mathcal{D} \cup \left(o_{t}, m_{t}^{\text{gt}}, a_{t}, r_{t}^{\text{env}}+ r_{t}^{\text{int}}, o_{t+1}\right)$

\State $\text{Update}\left(\mathcal{D}\right)$
    
\EndFor
\State \textbf{function} $\text{Update}\left(\mathcal{D}\right)$

\State \ \ \ \ $\left(o_{t}, m_{t}^{\text{gt}}, a_{t}, r_{t:t+n-1}, o_{t+n}\right) \sim \mathcal{D}$

\State \ \ \ \ Sample data augmentation $A$

\State \ \ \ \ $\hat{c}^{+}_{t}, \hat{c}^{-}_{t} \gets E_{\theta}\left(A\left(o_{t}\right)\right)$

\State \ \ \ \ $ m_{t}^{\text{pred}} \gets MP_{\phi} \left(\hat{c}^{+}_{t} \right)$

\State \ \ \ \ $\mathcal{L}_{\text{recons}} = \mathcal{L}_{\text{BCE}} \left(m_{t}^{\text{pred}}, A\left(m_{t}^{\text{gt}}\right)\right)$

\State \ \ \ \ Compute $\mathcal{L}_{\text{critic}}$ \Comment{See~\cite{DrQv2}}

\State \ \ \ \ $\mathcal{L}_{\text{total}} = \alpha \mathcal{L}_{\text{critic}} + \beta \mathcal{L}_{\text{recons}}$

\State \ \ \ \ Update Critic, $E_{\theta}$ and $MP_{\phi}$ using $\mathcal{L}_{\text{total}}$

\State \ \ \ \ Update Actor using RL

\State \textbf{end function}
\end{algorithmic}
\end{spacing}
\end{algorithm}

\section{Experiments}
\subsection{Environment}
\begin{figure}
    \centering \vspace{2mm}
    \includegraphics[width=1.0\linewidth]{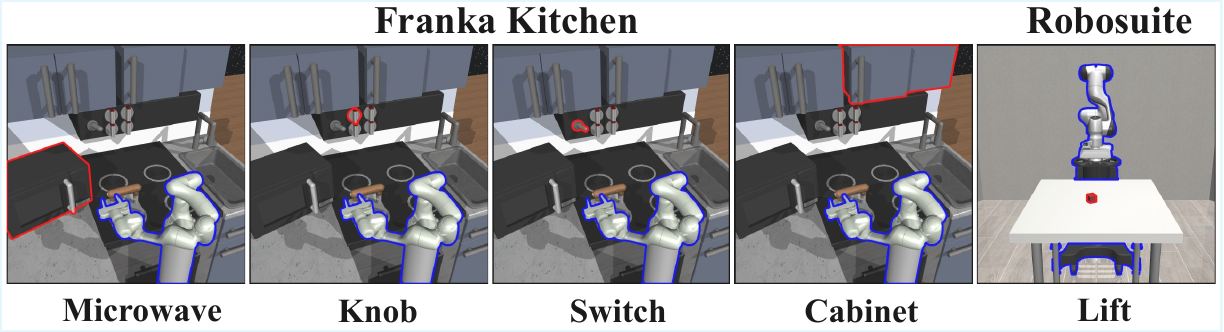}
    \caption{Simulation environments and tasks. We evaluate CDE on 5 challenging visual manipulation tasks.}
    \label{fig:environment}
\end{figure}

We conduct experiments to evaluate CDE's performance on two robot manipulation benchmarks (see \cref{fig:environment}).

\begin{figure}
    \centering
    \includegraphics[width=1.\linewidth]{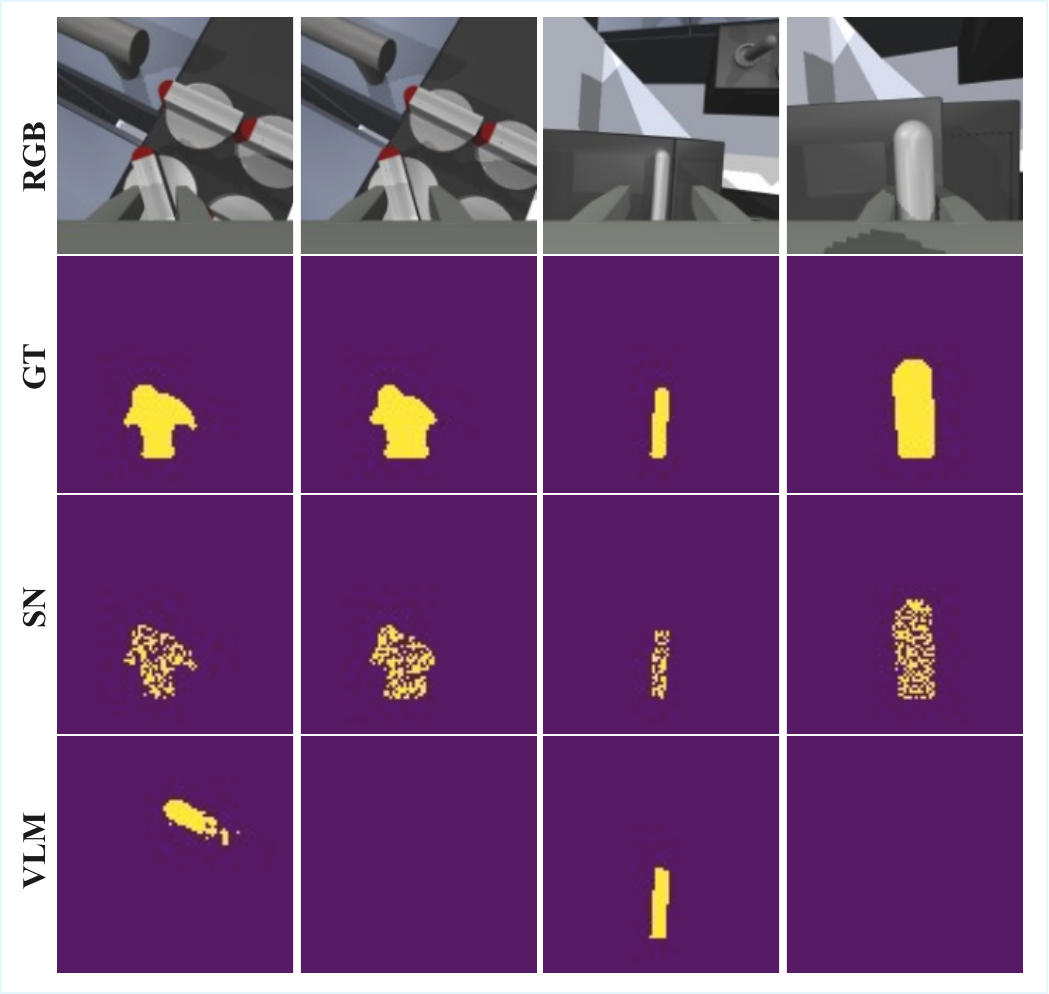}
    \caption{Mask settings. (Row 1) RGB input from the wrist-mounted camera. (Row 2) Ground truth (GT) masks acquired through the simulator. (Row 3) Synthetic Noise (SN) masks derived from GT masks via noise injection. (Row 4) VLM-generated masks.}
    \label{fig:mask examples}
\end{figure}

\textbf{Franka Kitchen}~\cite{franka_kitchen}: is a challenging RL benchmark for visual control. The agent needs to interact with various objects in the kitchen given only sparse reward signals. We choose 4 tasks: \textit{Microwave}, \textit{Knob}, \textit{ Switch} and \textit{Cabinet}.

\textbf{Robosuite}~\cite{robosuite}: contains several challenging robotic tabletop manipulation tasks. We evaluate on the \textit{Lift} task with sparse rewards.

\begin{figure*}
    \centering \vspace{2mm}
    \includegraphics[width=1.\linewidth]{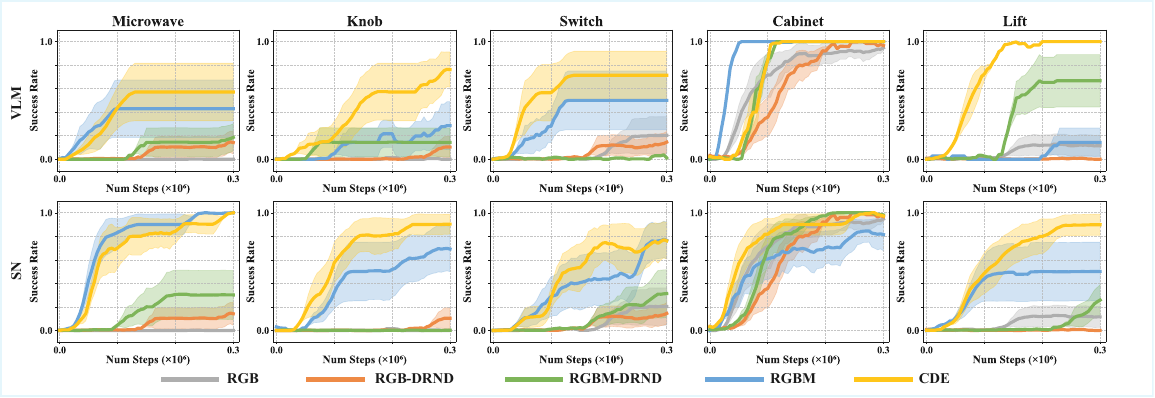}
    \caption{Training curves. RGB and RGB-DRND are trained without input segmentation masks, while RGBM-DRND, RGBM, and CDE leverage masks during training. We report average success rate with standard error across seven random seeds. (Top row) Learning with VLM-generated masks. (Bottom row) Learning with masks with synthetic noise. CDE achieves the highest average success rate on most scenarios while remaining more stable to noisy mask inputs than baselines.}
    \label{fig:training_curves}
\end{figure*}

\begin{figure*}
    \centering \vspace{2mm}
    \includegraphics[width=1.\linewidth]{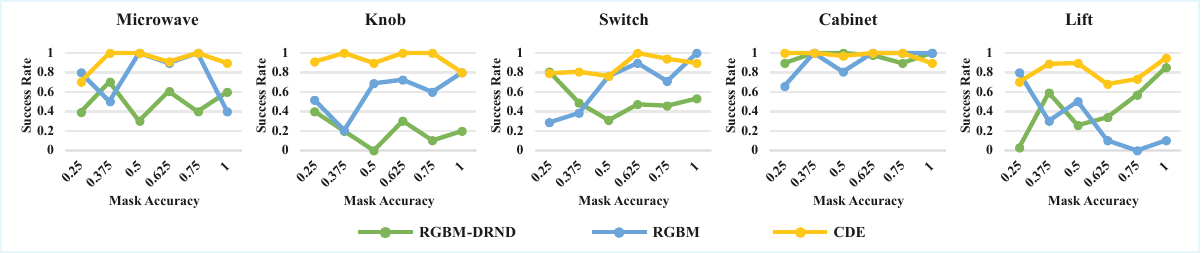}
    \caption{Performance under various levels of synthetic noise. All methods which leverage segmentation masks for policy learning are compared under six mask accuracy levels: 0, 0.25, 0.375, 0.50, 0.625, and 0.75. We report average success rate across ten random seeds for each noise level. CDE not only achieves high success rate on all tasks, but also shows strong robustness to noise level change.}
    \label{fig:synthetic_noise}
\end{figure*}

\subsection{Baselines}
We compare CDE with the following baselines, representing standard methods of leveraging RGB observations and incorporating segmentation masks into policy learning.

\begin{itemize}
    \item \textbf{RGB}: state-of-the-art model-free visual RL algorithm DrQv2~\cite{DrQv2} using RGB observations.

    \item \textbf{RGBM}: DrQv2 with segmentation mask concatenated as a fourth image channel.

    \item \textbf{RGB-DRND}: RGB baseline with additional intrinsic rewards provided by Distributional Random Network Distillation (DRND)~\cite{DRND}.

    \item \textbf{RGBM-DRND}: RGBM baseline with additional instrinsic rewards provided by DRND.

\end{itemize}

\subsection{Experimental Setup}
We evaluate CDE under three noise settings (See \cref{fig:mask examples}), including: ground truth (GT) masks, synthetic noise (SN) masks, and VLM-generated masks. GT masks are obtained through MuJoCo's~\cite{mujoco} rendering API. SN masks are derived from the GT mask by randomly inverting pixels according to a binomial distribution.
VLM masks are generated with Grounded-SAM2~\cite{groundedsam}, which integrates the open-vocabulary object detection model Grounding-DINO~\cite{groundingdino} with Segment Anything Model 2 (SAM2)~\cite{SAM2}. Notably, we only segment the mask on the first frame of the episode and use mask tracking to acquire masks for later frames.

For RGB baselines, we use the task reward only.
For RGBM, the segmentation mask is provided as input, so no reconstruction is available to supply an intrinsic reward.
Instead, we provide a shaping reward proportional to the number of pixels in the segmentation mask.
For RGB-DRND and RGBM-DRND, intrinsic reward is generated by prediction error between the target network and the predictor network given the observation.
For all environments, the resolution of both images and masks is $84\times 84$, and a frame stack of 3 is used. We run all the experiments for $3 \times 10^{5}$ time steps on an NVIDIA A100 GPU (40GB); each run takes around 4 hours with GT or SN mask, and around 8 hours with the VLM-generated mask.

\subsection{Results and Analysis}

\cref{fig:training_curves} shows the training curves for CDE compared to baseline methods when using both VLM-generated masks from Grounded-SAM2 (top row) as well as ground truth masks with additive synthetic noise (bottom row).
When utilizing VLM masks, we observe that CDE yields higher performance than all baseline methods, including those with intrinsic rewards (RGBM-DRND and RGB-DRND), in every task except for \textit{Cabinet}.
A similar pattern holds for the synthetic masks, where CDE outperforms all baselines except in \textit{Cabinet} and \textit{Microwave}.

We draw three conclusions from this.
First, because the RGBM baseline directly operates over masks, any errors present in the mask are propagated into both perception and exploration which degrades performance.
CDE, on the other hand, reconstructs masks from latent embeddings which reduces the impact of noise and results in better performance.
This is particularly important given the amount of error present in VLM-generated masks, as shown in Table~\ref{tab:IoU}.
Second, due to the challenging sparse reward setting of these tasks, RGB observations alone are insufficient to learn optimal policies, even when paired with an intrinsic reward (RGB-DRND).
This indicates that task-relevant information in the form of segmentation masks are necessary.
Third, intrinsic rewards as given by the baseline method DRND are surprisingly unhelpful, and even \textit{degrade} performance in many cases, e.g. RGBM-DRND in \textit{Switch} and \textit{Microwave}.

\begin{table}[b]
  \centering
  \small
  \setlength\tabcolsep{1.5pt}
  \resizebox{\linewidth}{!}{
  \begin{tabular}{r | *{4}{c} | *{5}{c} | c}
    \toprule

    & \multicolumn{4}{c|}{\textbf{Component}} & \multicolumn{5}{c}{\textbf{Success Rate $\left(\% \right)$}}\\
    \cline{2-11}
    & \rotatebox{0}{PE} & \rotatebox{0}{NE} & \rotatebox{0}{RR} & \rotatebox{0}{PR} & {Microwave} & {Knob} & {Switch} & {Cabinet} & {Lift} & {Average}\\
    \midrule
    \midrule
    1 \ \ & \checkmark & \checkmark & \checkmark  & \checkmark & $\hibf{90}$~\scriptsize{$\pm 09$} & 80~\scriptsize{$\pm 16$} & $\hibf{90}$~\scriptsize{$\pm 09$} & $\hibf{100}$~\scriptsize{$\pm 00$} & 40~\scriptsize{$\pm 23$} & 80 
    \\
    2 \ \ & \checkmark & \checkmark & \texttimes  & \checkmark & $\hibf{90}$~\scriptsize{$\pm 09$} & $\hibf{90}$~\scriptsize{$\pm 09$} & $\hibf{90}$~\scriptsize{$\pm 09$} & $\hibf{100}$~\scriptsize{$\pm 00$} & 48~\scriptsize{$\pm 24$} & 84 \\
    3 \ \ & \checkmark & \texttimes & \texttimes  & \checkmark & 80~\scriptsize{$\pm 16$} & $\hibf{90}$~\scriptsize{$\pm 09$} & 81~\scriptsize{$\pm 13$} & $\hibf{100}$~\scriptsize{$\pm 00$} & 40~\scriptsize{$\pm 24$} & 78
    \\
    4 \ \ & \checkmark & \texttimes & \checkmark  & \texttimes & 64~\scriptsize{$\pm 20$} & 60~\scriptsize{$\pm 24$} & 80~\scriptsize{$\pm 16$} & 87~\scriptsize{$\pm 09$} & 93~\scriptsize{$\pm 03$} & 77 \\
    5 \ \ & \checkmark & \checkmark & \texttimes  & \texttimes & 20~\scriptsize{$\pm 16$} & 31~\scriptsize{$\pm 20$} & 40~\scriptsize{$\pm 24$} & 99~\scriptsize{$\pm 01$} & 31~\scriptsize{$\pm 20$} & 44 \\
    \hline
    CDE & \checkmark & \checkmark & \checkmark  & \texttimes & $\hibf{90}$~\scriptsize{$\pm 09$} & 77~\scriptsize{$\pm 16$} & $\hibf{90}$~\scriptsize{$\pm 09$} & 90~\scriptsize{$\pm 09$} & $\hibf{95}$~\scriptsize{$\pm 02$} & $\hibf{88}$ \\
    \bottomrule
  \end{tabular}
  }
  \caption{Ablation studies. PE stands for positive embedding, NE stands for negative embedding, RR stands for reconstruction reward and PR stands for pixel reward. We report the average success rate with standard error, the highest success rate for each task is highlighted in yellow.}
  \label{tab:concept-ablation}
\end{table}

\begin{figure*}[!t]
    \centering \vspace{2mm}
    \includegraphics[width=0.9\linewidth]{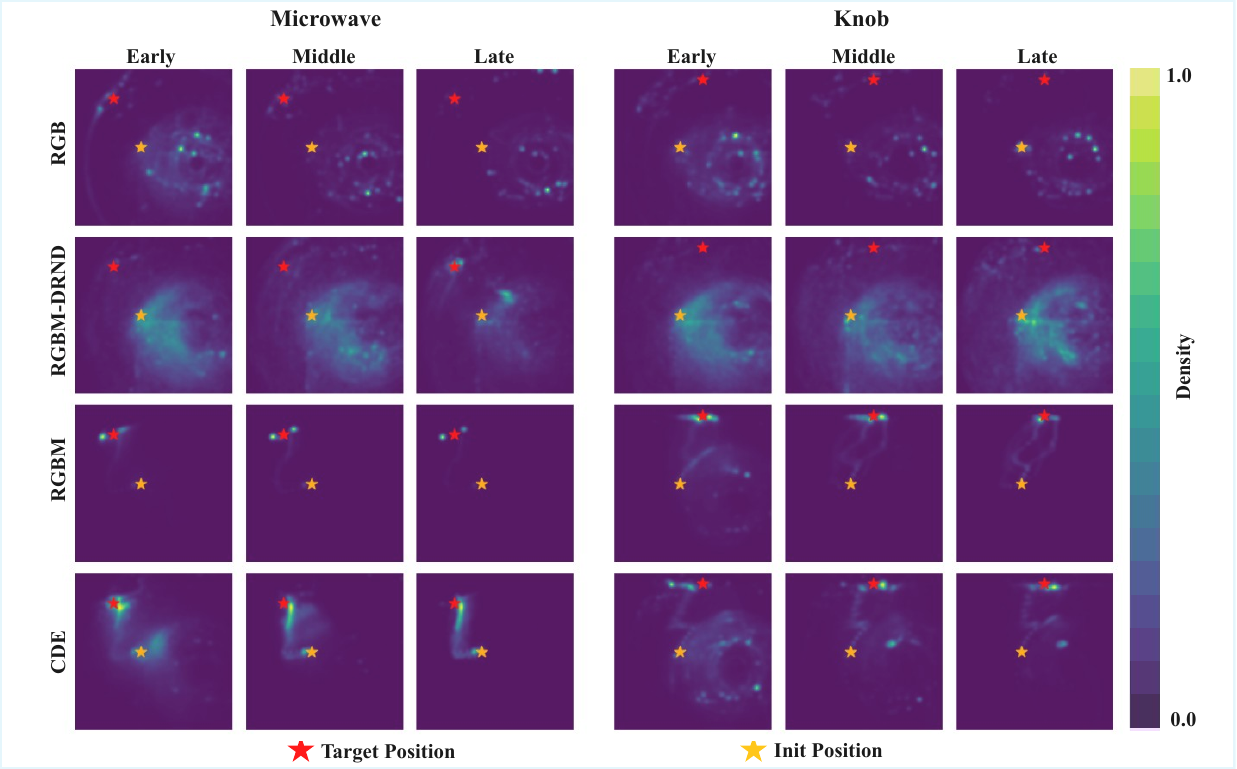}
    \caption{Heatmap showing visited states during three different stages of training: Early (33\%), Middle (66\%), Late (100\%). RGB explores largely randomly. RGBM-DRND encourages searching for unseen states, achieving high state coverage. RGBM concentrates near the target object to maximize pixel-based reward. CDE exhibits intelligent exploration strategy by effectively interacting with the target object.}
    \label{fig:Exploration Heatmap}
\end{figure*}

\cref{fig:synthetic_noise} shows the performance impact of CDE and RGBM-based baseline methods under varying levels of synthetically-induced mask errors.
These results show that CDE yields strong performance across a wide range of noise levels, exhibiting \textit{at least} a 70\% success rate for every task, even when the input masks are only 25\% accurate.
In contrast, RGBM and RGBM-DRND exhibit far higher levels of performance degradation as mask errors increase.
RGBM sees success rates as low as 30\% (Switch, 0.25) while RGBM-DRND collapses completely to 0\% (Lift, 0.25).
These findings support our claim that CDE mitigates the negative performance impact of VLM errors by treating segmentation masks as weak supervisory signals, resulting in performance that is nearly invariant to mask errors.

\begin{table}
\centering
\small
  \resizebox{\linewidth}{!}{
 
\begin{tabular}{r c c c c c }    
\toprule & \multicolumn{5}{c}{\textbf{Task}} \\
 \cmidrule {2-6} 
 \textbf{Metric} & Microwave & Knob & Switch & Cabinet & Lift \\ 
 \midrule  
 IoU & 0.228 & 0.007 & 0.657 & 0.170 & 0.910  \\  
 \bottomrule   

\end{tabular}  
}
  \caption{Average IoU between GT mask and VLM-generated mask.}
  \label{tab:IoU}
\end{table}

\subsection{Analysis of Mask Errors}
To further analyze the impact of VLM-generated mask errors on the performance, we collect a dataset containing VLM-generated masks and GT masks across 5 tasks. Specifically, we train the policy with GT masks for $1 \times 10^{5}$ time steps and execute 5 policy rollout every $1 \times 10^{4}$ time steps to collect RGB observations and ground truth masks, resulting in 50 trajectories for each task. We then employ Grounded-SAM2 on the collected dataset to acquire VLM-generated masks for each single RGB image. To quantify the quality of VLM-generated masks, we compute Intersection over Union (IoU) for all GT-VLM mask pairs on each task and report average IoU as shown in \cref{tab:IoU}. We observe that the average IoU on the \textit{Knob} task is extremely low, indicating that the VLM produces erroneous predictions.
This can be seen in \cref{fig:mask examples} where the VLM segments the wrong mask (first column on the left) or fails to identify it completely (second column).
Additionally, we note that despite the poor accuracy of VLM-generated masks, all methods achieve strong performance on the \textit{Cabinet} task. 
We attribute this to the fact that \textit{Cabinet} requires fewer interactions with the target object to achieve success compared to other tasks such as \textit{Microwave}.

\subsection{Ablation Studies}
\label{sec:ablation studies}
We conduct ablation studies to investigate the contribution of key components of CDE (See \cref{tab:concept-ablation}). 
First, using both positive and negative embeddings (CDE, Model 2) outperforms variants with only positive embeddings (Model 4, Model 3).
This confirms that the CEM helps the policy in cases where the object is not visible, improving learning.

Second, we compare the impact of pixel reward (PR) and reconstruction reward (RR). Although variants with PR (Model 2) and PR$+$RR (Model 1) show strong performance on Franka Kitchen tasks, their performance degrades significantly on the \textit{Lift} task, whereas our RR-based method remains stable across all tasks and achieves the highest average success rate.
This suggests that PR is sensitive to fine-grained mask accuracy and visual noise, while RR provides a more robust, task-agnostic intrinsic signal. 

Additionally, we also observe that CDE outperforms its variant without PR or RR (Model 5) on most tasks, this indicates that RR provides object-centric reward signals, encouraging exploration. However, in some simple scenarios such as the \textit{Cabinet} task which do not require many interactions, such exploration may decrease the sample efficiency.

\subsection{Exploration Analysis}
To better understand how CDE influences exploration, we record end-effector positions at early, middle, and late stages of training and visualize the positions in a heatmap as shown in \cref{fig:Exploration Heatmap}. In the early stage, all methods begin exploring the environment. However, their positional distributions reveal four distinct exploration strategies.

For RGB, the heatmap exhibits several high-density regions near the initial position. Since exploration for the policy is completely random and constrained by robot arm's kinematics due to sparse environment reward, the trajectories remain concentrated around the initial position. For RGBM-DRND, the exploration trajectories are more uniformly distributed without pronounced high-density regions. The intrinsic reward generation of RND encourages visitation of novel states, resulting in a more even spatial coverage.

Unlike previous methods that largely rely on non-object-centric exploration, RGBM is instead guided by a direct pixel-based shaping reward, making the trajectory primarily concentrated near the target object.  However, such direct reward signals can also compromise exploration; the policy opts to maximize the size of segmentation mask to receive higher pixel-based reward rather than effectively interacting with the object. In contrast, CDE exhibits exploration around both initial position and target position, where the target object remains visible. This visibility enables the policy to learn visual representations and generate reconstruction that guide exploration.

At the middle stage, the three baseline methods largely maintain spatial patterns similar to those observed before. In comparison, CDE develops a high-density region around the target object, indicating that the policy has identified the object and begins consistent interaction with it. By the late stage, CDE consistently completes the task and the heatmap exhibits a concentrated trajectory around the learned solution.

\begin{figure}
    \centering
    \begin{subfigure}[b]{0.6\linewidth}
    \includegraphics[width=\linewidth, height=2.9cm]{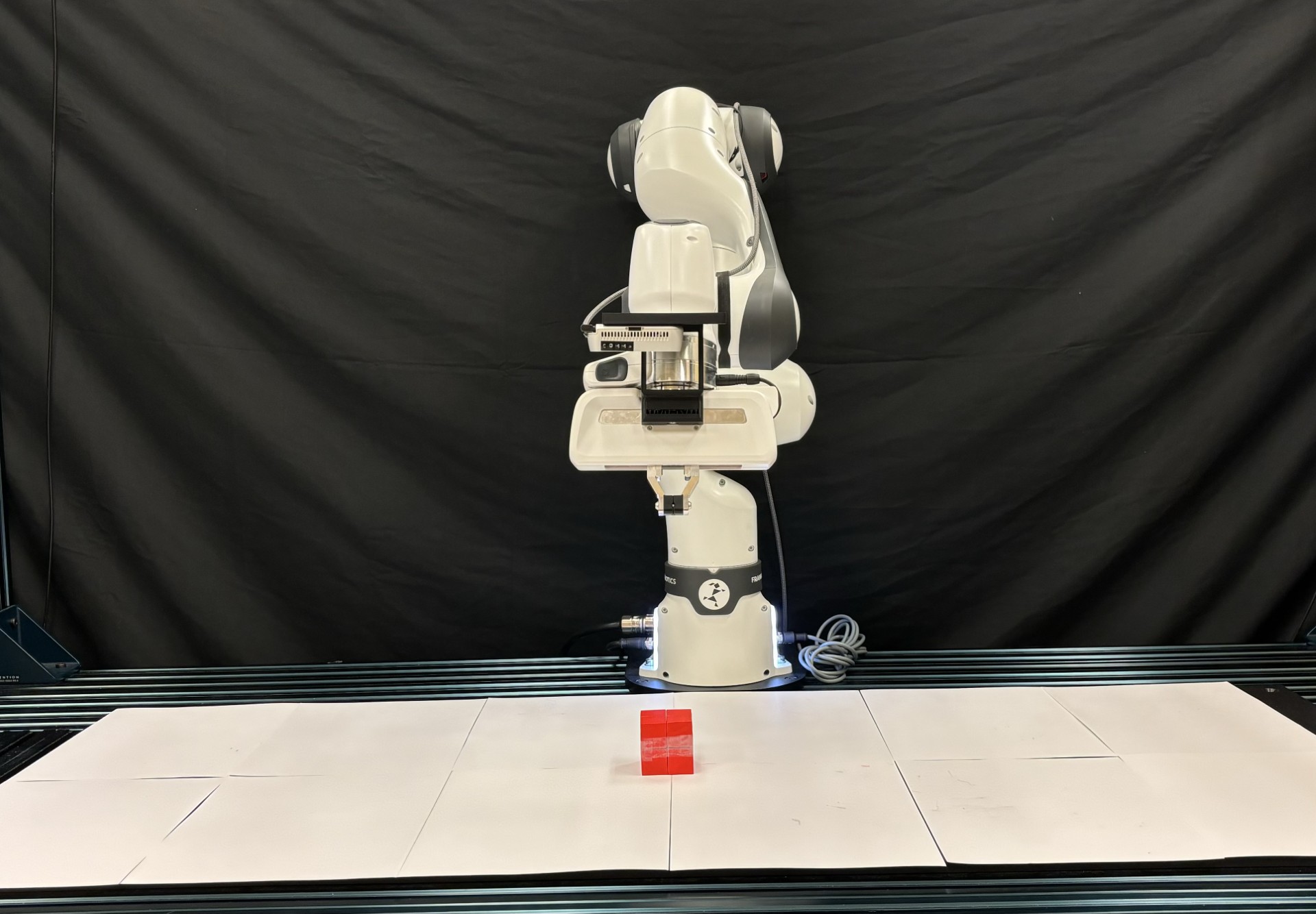}
    \caption{}
    \label{fig:setup}
    \end{subfigure}
    \begin{subfigure}[b]{0.3\linewidth}
    \includegraphics[width=\linewidth, height=2.9cm]{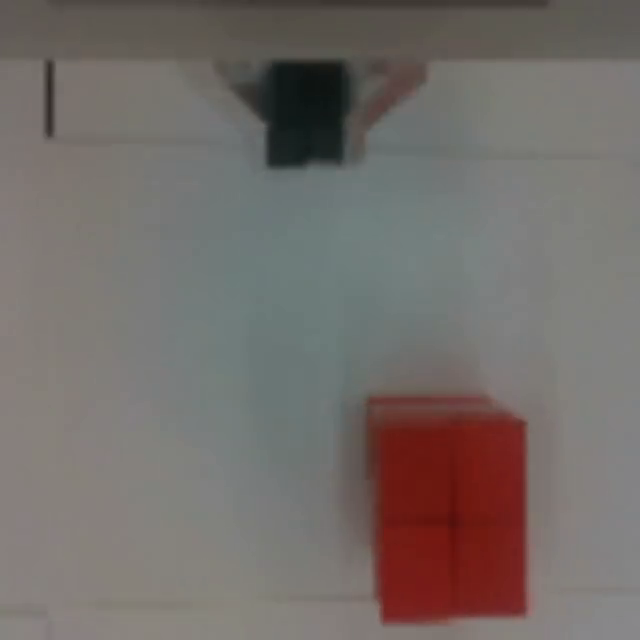}
    \caption{}
    \label{fig:real world image input}
    \end{subfigure}
    \caption{(a) Real-world experiment setup. (b) Wrist-mounted camera view.}
    \label{fig:real_world}
\end{figure}

\subsection{Real-World Experiments}
We also conduct real-world experiments with a Franka Research 3 robot arm and an Intel Realsense D435i camera mounted on the wrist as shown in \cref{fig:setup}. We train a policy on the \textit{Lift} task with an end-effector position controller.
We directly perform sim-to-real transfer without any fine-tuning for CDE.
We execute 10 policy rollouts, resulting in 8/10 successes for CDE. This result demonstrates CDE's strong capability of real world deployment.

\section{Conclusion}
In this paper, we propose CDE, a novel concept-driven exploration method for RL, which uses a VLM to discover task-relevant visual concepts.
CDE treats VLM outputs as noisy supervision for representation learning and uses reconstruction errors as intrinsic rewards, yielding generalizable object-centric exploration without relying on VLM inputs at test-time.
Our approach supports wrist-mounted camera observations by learning dual object representations: one embedding when the object is visible, and one when it is not.
This allows the policy to learn complementary features for each behavior mode, i.e. searching for the object vs interacting with it.
Our experiments show that CDE is more stable and robust compared to baselines in both simulated and real-world manipulation tasks.
CDE is simple to plug into existing RL frameworks, and we believe it opens avenues for further research in efficient object-centric exploration.

\bibliographystyle{IEEEtran}{
\bibliography{refs}}
\addtolength{\textheight}{-12cm} 
\end{document}